\documentclass[conference]{IEEEtran}
\IEEEoverridecommandlockouts

\usepackage{algorithmic}
\usepackage{array}
\usepackage{textcomp}
\usepackage{stfloats}
\usepackage{url}
\usepackage{graphicx}
\usepackage{amsmath,amssymb,amsfonts}
\usepackage{multirow,multicol}
\usepackage{todonotes}
\usepackage{microtype}
\usepackage{cleveref}
\usepackage{microtype}
\newcommand{\tip}{\mathrm{tip}}
\usepackage{microtype}

\usepackage{overpic}
\usepackage{booktabs}
\usepackage{colortbl}
\usepackage{fontawesome5}
\usepackage{pgf-pie} 

\usepackage{caption}
\usepackage{cite}
\usepackage{amsmath,amssymb,amsfonts}
\usepackage{algorithmic}
\usepackage{graphicx}
\usepackage{textcomp}
\usepackage{xcolor}
\usepackage{xcolor}
\definecolor{cvprblue}{RGB}{0,114,189}

\usepackage{pgfplots}
\pgfplotsset{compat=1.18}
\usepackage{pgf-pie}
\definecolor{cMisc}{HTML}{B7B7B0}   

\definecolor{cEuler}{HTML}{D98B9F}  
\definecolor{cDino}{HTML}{8E8AD6}   
\definecolor{cRansac}{HTML}{6FC3B8} 
\definecolor{cDit}{HTML}{F0A860}    
\definecolor{cIcp}{HTML}{E8D26A}    
\definecolor{cDgedi}{HTML}{7CB66B}  
\definecolor{cUmap}{HTML}{E79CCB}   

\begin{document}

\title{Training-Free Hold-Usage Detection in Sport Climbing with Foundation Pose Models}
 
\author{\IEEEauthorblockN{Abu Bakar}
\IEEEauthorblockA{\textit{Virtual University of Pakistan} \\
Islamabad, Pakistan\\
chbakar26@gmail.com}
\and
\IEEEauthorblockN{Abdullah Aftab}
\IEEEauthorblockA{\textit{Army Public School and Colleges} \\
Rawalpindi, Pakistan\\
abdullahaftabaes98@gmail.com}
\and
\IEEEauthorblockN{Amir Hamza}
\IEEEauthorblockA{\textit{University of Trento, Italy} \\
Trento, Italy \\
amir.hamza@unitn.it}
}

\maketitle

\IEEEpubid{\makebox[\columnwidth]{%
    \fontsize{7}{8}\selectfont
    979-8-3195-0247-6/26/\$31.00~\copyright~2026 IEEE\hfill}%
    \hspace{\columnsep}\makebox[\columnwidth]{}}

\IEEEpubidadjcol

\begin{abstract}
Detecting which holds a climber uses, and when, underpins automated scoring, movement analysis, and assistive systems for sport climbing. Existing approaches train task-specific models or repurpose 2D pose estimators whose hand keypoint sits at the wrist and foot keypoint at the ankle i.e. offset from the fingertips and toes that actually contact the holds, and whose hands are occluded in roughly half of all frames. We show that a \emph{frozen, off-the-shelf} pose foundation model is sufficient: using the fingertip and toe keypoints of Sapiens, a per-frame proximity test against the annotated holds, per-limb mutual exclusion, and a short temporal-persistence rule, we detect hold usage \emph{without any climbing-specific training}. On the \emph{The Way Up} dataset (22 videos, 10 athletes, two routes), our method reaches an event $F_1$ of $90.2\%$ on a held-out split ($89.8\%$ under leave-one-participant-out cross-validation) and $79.9\%$ over all 22 videos at any temporal overlap, and performs best on footholds ($F_1\,89.8\%$ overall, $96.6\%$ held-out). Under an identical protocol it exceeds our reproductions of the YOLOv8-pose and ViTPose pipelines at every temporal threshold, with the margin widening under strict timing. An ablation shows that two intuitively helpful additions---dense foundation-feature change gating and body-part segmentation---both \emph{hurt}, arguing that a minimal, keypoint-only design is the right one for this task. Finally, standard coaching statistics computed from our automatic predictions track ground truth closely (Pearson $r=1.00$ for climb time, $0.94$ for pace), turning ordinary single-camera video into reliable performance metrics with no instrumentation.
\end{abstract}


\section{Introduction}
\label{sec:intro}
 
Climbing has grown rapidly, from its Olympic debut and expansion to the proliferation of indoor gyms and youth programs. This momentum has created real demand for tools that analyze climbs automatically for scoring competitions, giving athletes technique feedback, generating gym analytics, and powering assistive systems for visually impaired climbers~\cite{michenthaler2022scoring, boulanger2016sensor, richardson2022climbovision}. Almost all of these applications rest on the same primitive: knowing \emph{which holds an athlete uses, and when}.

Today, this information is obtained manually or via instrumented walls, neither of which scales. Maschek et al. formalized a video-based alternative with \emph{The Way Up} dataset, casting hold-usage detection as a 2D-pose problem~\cite{maschek2025wayup}. Their baseline uses wrist and ankle keypoints, requiring an inflated region of interest to overlap holds. This inherits two weaknesses: a \emph{localisation} problem (climbers contact holds at fingertips and toes, not wrists and ankles) and an \emph{occlusion} problem (hands disappear in roughly half of all annotated frames, degrading the wrist signal). 
We show a frozen, off-the-shelf human-pose foundation model addresses both problems without climbing-specific training. Sapiens~\cite{khirodkar2024sapiens} natively predicts 308 keypoints, including individual fingers and toes. By taking these true contact keypoints directly, scoring their proximity to annotated holds, resolving per-limb ties, and applying a temporal-persistence rule, we detect hold usage accurately. The fingertip and toe keypoints make inflated regions unnecessary, and the temporal pass bridges brief occlusions. Counterintuitively, adding richer foundation cues, such as dense feature-change gating or body-part segmentation, actually degrades performance. The dominant error source is keypoint localization, rewarding a minimal, keypoint-only design.

We evaluate on \emph{The Way Up} (22 videos, 10 athletes, two routes) using the dataset's task and temporal-IoU metric. Our contributions are:
\begin{itemize}
  \item A \textbf{training-free} hold-usage detector built from frozen foundation models (a promptable segmenter~\cite{kirillov2023sam, ravi2025sam2} for the climber box and Sapiens~\cite{khirodkar2024sapiens} for pose), generalizing to unseen environments.
  \item A \textbf{same-protocol comparison} against YOLOv8-pose and ViTPose pipelines, and an \textbf{ablation} showing that feature gating and body-part segmentation degrade performance while minimalism wins.
  \item An \textbf{occlusion-stratified} evaluation quantifying performance in heavily occluded scenarios, which prior work highlights as hardest but does not measure.
\end{itemize}

\section{Related Work}
\label{sec:related}
 
\noindent\textbf{Vision for climbing.}
Camera-based systems analyze climbing using custom detectors or generic pose models for competition scoring, technique evaluation, and speed climbing~\cite{michenthaler2022scoring, vrzakova2024climbing, beltran2023technique, cao2021openpose, pandurevic2022speed, elias2021speed21}. Reviews of sensor and camera methods~\cite{andric2022review, boulanger2016climbing} show that prior works either train task-specific detectors or use generic wrist and ankle keypoints. None exploit dense fingertip or toe keypoints, and none operate training-free.
 
\noindent\textbf{Pose-based hold-usage detection.}
Closest to our approach, Maschek and Schedl~\cite{maschek2025wayup} introduced \emph{The Way Up} and evaluated baseline models~\cite{lugaresi2019mediapipe, xu2022vitpose, jocher2023yolov8} that use wrist and ankle keypoints to define an area of interest, triggering usage after $0.5$\,s of overlap. Because ankle-only models place the keypoint far from the toe, they require large and imprecise foot regions, which the authors identify as a major error source. We adopt their dataset and metric but replace wrist and ankle keypoints with fingertips and toes, add per-limb mutual exclusion, and require no training.
 
\noindent\textbf{Foundation models.}
Promptable segmentation~\cite{kirillov2023sam, ravi2024sam2} and self-supervised features~\cite{oquab2024dinov2} generalize zero-shot across domains. Sapiens~\cite{khirodkar2024sapiens} natively predicts $308$ keypoints, including the fingers and toes absent from COCO-style skeletons. We use these models frozen. A promptable segmenter supplies the climber box and Sapiens supplies the contact keypoints with no climbing adaptation.
 
\noindent\textbf{Climbing datasets.}
Prior datasets target 3D motion, gaze, or speed climbing~\cite{yan2023cimi4d, vrzakova2024climbing, elias2021speed21}. \emph{The Way Up}~\cite{maschek2025wayup} is the only dataset annotating which limb uses which hold over specific frame intervals, making it the natural benchmark for our task.

\section{Method}
\label{sec:method}

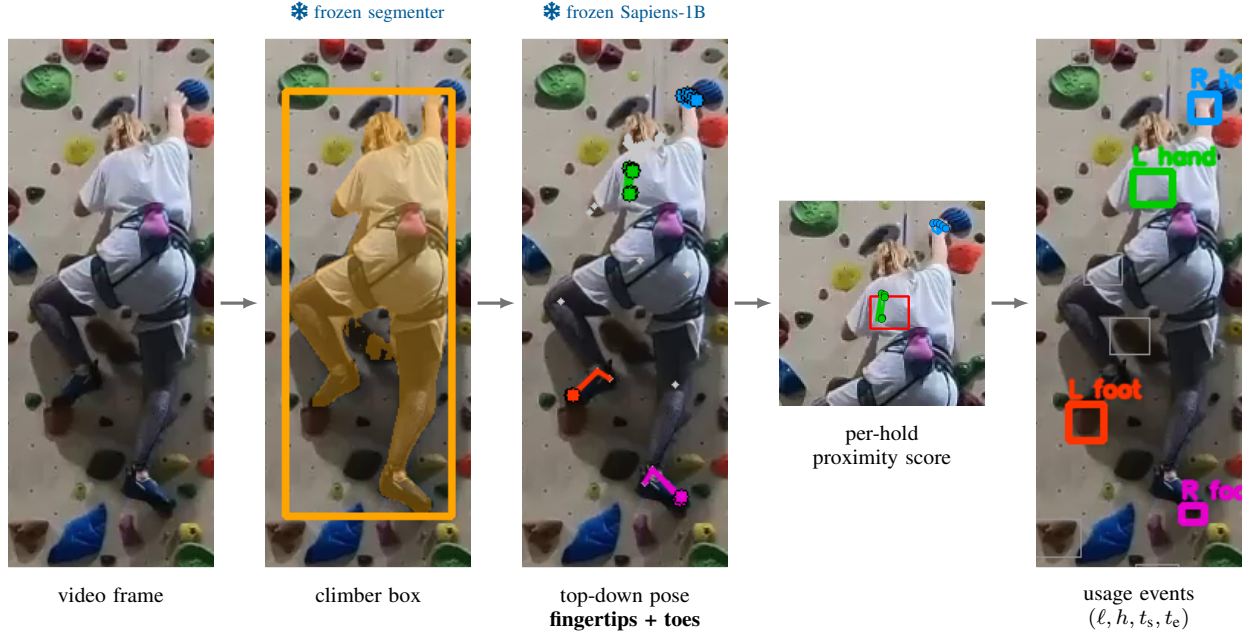
\begin{figure*}[t]\centering
\usetikzlibrary{arrows.meta,positioning}
\begin{tikzpicture}[font=\scriptsize,
  panel/.style={inner sep=0pt, outer sep=0pt, draw=black!25, line width=0.4pt},
  lbl/.style={align=center, font=\footnotesize, text width=2.7cm},
  tag/.style={align=center, font=\scriptsize, text=cvprblue!80!black},
  ar/.style={-{Latex[length=2mm]}, very thick, black!55}]

  \node[panel] (p1) {\includegraphics[width=2.7cm]{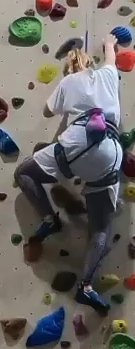}};
  \node[panel, right=7mm of p1] (p2) {\includegraphics[width=2.7cm]{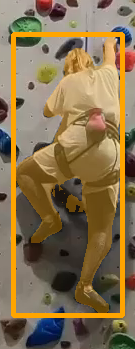}};
  \node[panel, right=7mm of p2] (p3) {\includegraphics[width=2.7cm]{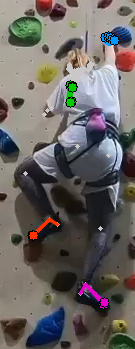}};
  \node[panel, right=7mm of p3] (p4) {\includegraphics[width=2.7cm]{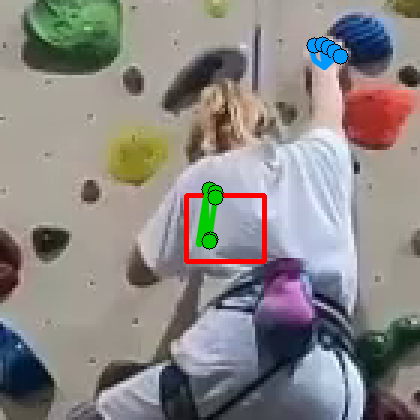}};
  \node[panel, right=7mm of p4] (p5) {\includegraphics[width=2.7cm]{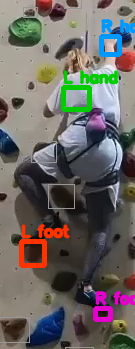}};


    \draw[ar, shorten >=1mm, shorten <=1mm] (p1.east) -- (p2.west);
    \draw[ar, shorten >=1mm, shorten <=1mm] (p2.east) -- (p3.west);
    \draw[ar, shorten >=1mm, shorten <=1mm] (p3.east) -- (p4.west);
    \draw[ar, shorten >=1mm, shorten <=1mm] (p4.east) -- (p5.west);
    
  \node[lbl, below=1.5mm of p1] {video frame};
  \node[lbl, below=1.5mm of p2] {climber box};
  \node[lbl, below=1.5mm of p3] {top-down pose\\\textbf{fingertips + toes}};
  \node[lbl, below=1.5mm of p4] {per-hold\\proximity score};
  \node[lbl, below=1.5mm of p5] {usage events\\$(\ell,h,t_\text{s},t_\text{e})$};

  \node[tag, above=1mm of p2] {\faSnowflake\ frozen segmenter};
  \node[tag, above=1mm of p3] {\faSnowflake\ frozen Sapiens-1B};

\end{tikzpicture}
\caption{\textbf{Training-free hold-usage pipeline.} Snowflakes (\faSnowflake) mark
the frozen, off-the-shelf foundation models, which receive no climbing-specific
training. A promptable segmenter~\cite{kirillov2023sam} returns the climber box,
Sapiens~\cite{khirodkar2024sapiens} estimates top-down pose with 308 whole-body
keypoints of which we keep the fingertips and toes, and the proximity of each tip
to the given hold box (\cref{eq:prox}), after per-limb mutual exclusion and a
 short temporal pass, yields the usage events.}
\label{fig:pipeline}
\end{figure*}

We detect hold usage training-free using frozen foundation models. The input is a video and hold bounding boxes; the output is a set of usage events defining the limb, hold, and frame interval. Fig. \ref{fig:pipeline} shows the pipeline. We process every second frame to keep inference cheap, relying on our temporal pass to bridge any short gaps.

\noindent\textbf{Climber localization.}
We run a promptable segmenter~\cite{kirillov2023sam} with a \emph{person} prompt, keeping the largest mask's bounding box. This confines pose estimation to the athlete, eliminating background distractions and bystander detections. Any zero-shot detector could serve this role.

\noindent\textbf{Fingertip and toe pose.}
Sapiens-1B~\cite{khirodkar2024sapiens} runs top-down on the climber box, predicting $308$ keypoints. We keep only the true contact points: the fingertip of each hand and the big toe tip of each foot. Reading these directly is our central design choice. Earlier works use wrists and ankles, which sit far from the contact point and require an enlarged, imprecise region of interest. Using fingertips and toes removes this offset entirely, making such regions unnecessary.

\noindent\textbf{Per-frame contact score.}
For limb $\ell$ and hold $h$ at frame $t$, we convert their distance into a score in $[0,1]$:
\begin{equation}
  s_\ell^h(t) = \operatorname{clip}\!\Big(1 - \tfrac{d(\tip_\ell(t),\, B_h)}{\tau},\; 0,\; 1\Big),
  \label{eq:prox}
\end{equation}
where $\tip_\ell(t)$ is the tip keypoint, $B_h$ the hold box, $d(\cdot,\cdot)$ the point-to-box distance, and $\tau$ a margin in pixels. The score is $1$ inside the hold and decays to $0$ at distance $\tau$, controlling forgiveness for localization errors. A per-frame mutual exclusion rule assigns each limb only its highest-scoring hold, removing spurious ties on densely packed routes.

\noindent\textbf{Temporal aggregation.}
Single-frame decisions are noisy due to pose jitter, brief occlusions, or hands brushing past holds. We convert the sequence $\{s_\ell^h(t)\}_t$ into clean intervals via a four-stage temporal pass: (1) \emph{smoothing} with a moving average to suppress spikes; (2) \emph{hysteresis} with separate enter and exit thresholds to prevent boundary toggling; (3) \emph{gap bridging} to merge intervals separated by few frames, preventing occlusions from splitting a grasp; and (4) \emph{dwell filtering} to discard intervals shorter than a threshold, removing transient contacts~\cite{maschek2025wayup}. Surviving intervals become our predicted usages $(\ell, h, t_\text{start}, t_\text{end})$. This linear-time pass costs almost nothing compared to pose inference.

\noindent\textbf{Minimalism over richer cues.}
We tested two appearance-based cues: an occlusion-robust gate using dense feature changes against an empty-wall clean plate~\cite{oquab2024dinov2}, and an overlap test using Sapiens' body-part segmentation. Both lowered accuracy (\cref{tab:abl}). The task rewards a minimal design because keypoint localization, not appearance or semantics, is the dominant source of uncertainty. Adding appearance modelling introduces new failure modes (e.g., suppressing real contacts if visual changes are subtle, or failing at image scales where hands vanish) without solving the localization bottleneck. Therefore, our final model relies strictly on keypoint proximity.

\section{Evaluation Protocol}
\label{sec:eval}

\noindent\textbf{Data and task.}
We evaluate on \emph{The Way Up}~\cite{maschek2025wayup}. The
dataset contains $22$ videos, formed from $11$ participant recordings of $10$
athletes climbing two routes, an easier ``orange'' route graded 4c and a harder
``green'' route graded 5a, on a vertical indoor wall. The footage was captured at
$50$\,FPS and is provided at a resolution of $720\times1280$ and a frame rate of
$25$\,FPS. Every video comes with two kinds of annotation. The first is a
bounding box for each hold on the route. The second is a list of ground truth
usages, where each usage records the limb involved, the hold it used, the start
and end frames of the contact, and any frame ranges in which that limb is more
than $50\%$ occluded. We take the provided hold boxes as given and our task is to
detect the usages.

\noindent\textbf{Split and hyperparameters.}
To be sure we never tune on the data we report on, we set aside participant
\texttt{p1}, across both routes, as a dedicated \emph{validation} split.
Every hyperparameter, ours as well as every baseline's, is chosen on \texttt{p1} alone
and then frozen, and we report results on the remaining $20$ videos. The
selection criterion is the mean $F_1$ over the three temporal IoU thresholds
defined below, again measured on \texttt{p1}. The frozen values for our method
are a margin of $\tau=20$\,px, a dwell threshold of $20$ frames, and hysteresis
enter and exit levels of $0.4$ and $0.15$.

\noindent\textbf{Protocols.}
Reporting a single split invites two opposite objections. A held out split can be
accused of wasting most of an already small dataset, while tuning on all of the
data, even without training, risks picking parameters that happen to flatter the
test videos. To answer both at once, we report three protocols and show that they
agree. The first is \emph{Held out 20}, where we tune on \texttt{p1}, freeze, and
test on the other $20$ videos. This is the clean generalization number and we
treat it as our primary result. The second is \emph{LOPO}, a leave one
participant out cross validation in which we tune on all participants but one and
test on the participant left out, repeating this so that every video is tested
exactly once with parameters that were never fit to it. This is the most rigorous
way to use the full dataset. The third is \emph{All 22, \texttt{p1} fixed}, where
we simply apply the \texttt{p1} frozen parameters to all $22$ videos. This is a
convenient full dataset number, though it is mildly optimistic on \texttt{p1}
since those videos helped choose the parameters. The three protocols agree to
within $0.4$ $F_1$ (\cref{tab:protocol}), which indicates that nothing is overfit
to the validation participant. We therefore use the held out number as the
headline result and the all $22$ number (\cref{tab:main}) for the finer grained
breakdown by route and limb.

\noindent\textbf{Metrics.}
We score at the level of usage \emph{events} rather than individual frames. A
predicted usage and a ground truth usage are eligible to match only when they
involve the same limb and the same hold. Among eligible pairs, we measure
temporal overlap with the temporal Intersection over Union,
\begin{equation}
  \operatorname{tIoU} = \frac{|T_p \cap T_g|}{|T_p \cup T_g|},
  \label{eq:tiou}
\end{equation}
where $T_p$ and $T_g$ are the predicted and ground truth frame
intervals~\cite{caba2015activitynet}. Matching is greedy and ordered by score. We
sort the predicted events by confidence and assign each one to the unmatched
ground truth event of the same limb and hold with which it has the highest tIoU,
provided that overlap clears the threshold, so that every event is used at most
once. A predicted event that finds a match is a true positive, a predicted event
that does not is a false positive, and a ground truth usage left unmatched is a
false negative. We exclude true negatives, because the vast majority of holds in
any video are never used and counting them would inflate every
score~\cite{maschek2025wayup}. From these counts we report Precision, defined as
$\mathrm{TP}/(\mathrm{TP}+\mathrm{FP})$, Recall, also called sensitivity and
defined as $\mathrm{TP}/(\mathrm{TP}+\mathrm{FN})$, and their harmonic mean
$F_1$. We report all three at a tIoU above $0$, meaning any temporal overlap at
all, which is the dataset paper's primary setting, and also at the stricter
thresholds of $0.3$ and $0.5$, together with the mean tIoU over matched events as
a direct measure of timing quality. We give these metrics overall, split into
hands and feet, and split by route, and we macro average over videos so that long
climbs do not dominate the score. Finally we report \emph{occlusion stratified
frame recall}, that is, recall computed separately over the ground truth frames
flagged as occluded and those flagged as visible, which directly quantifies
performance in the regime the dataset identifies as hardest but never measures.

\noindent\textbf{Baselines.}
We reproduce the dataset paper's keypoint method under our own evaluation so that
the comparison is fair. In that method, each model's wrist or ankle keypoint
defines an area of interest, which is a box centered on the wrist for hands and a
box extended downward for feet, made larger when only the ankle keypoint is
available. The frame is then cropped to the route region, keypoints are stored
across frames so that a momentary detection failure does not break tracking, and
a hold is counted as used once the area of interest overlaps it for at least
$0.5$\,s. We use the same two backbones as the dataset paper, namely
YOLOv8-pose~\cite{jocher2023yolov8} with its ankle keypoints and
ViTPose-L~\cite{xu2022vitpose} with its coco\_25 toe keypoints, and we tune each
baseline's area of interest margin and overlap thresholds on \texttt{p1} under
exactly the procedure described above. Our reproductions land below the numbers
published in the dataset paper, because we do not replicate their exact per model
margin tuning. The comparison we draw is therefore a \emph{same conditions} one,
in which our method and the baselines are evaluated identically, and we cite the
published numbers only for context (\cref{tab:cmp}).

\section{Results}
\label{sec:results}

Across all $22$ videos at $\operatorname{tIoU}>0$, our pipeline reaches an event
$F_1$ of $79.9\%$, with a Precision of $68.5$ and a Recall of $97.2$, and it
reaches $90.2\%$ on the held out $20$ videos under the frozen \texttt{p1}
parameters. 
The accuracy breaks down sharply by limb. Footholds are detected best,
at an $F_1$ of $89.8\%$ over all videos and $96.6\%$ on the held out split, while
handholds are lower but still strong at $72.7\%$. This pattern mirrors the
hand and foot gap reported for prior models, and it largely closes it, because
the toe keypoints localize feet very precisely whereas hands stay harder on
account of occlusion.
The route comparison follows the same logic as the dataset paper, where the easier green route is detected better than the orange one, at an
$F_1$ of $84.9$ against $74.9$ overall, the orange route being harder because its
holds sit close together.
Recall stays very high throughout, at $97.2\%$ overall and $98.9\%$ for feet, so the limiting factor for our method is over prediction.

\begin{table}[htp]

\centering
\footnotesize
\setlength{\tabcolsep}{2pt}

\resizebox{\columnwidth}{!}{%
\begin{tabular}{llccccccccc}
\toprule
& & \multicolumn{3}{c}{orange (4c)} & \multicolumn{3}{c}{green (5a)} & \multicolumn{3}{c}{both ($n{=}22$)}\\
\cmidrule(lr){3-5}\cmidrule(lr){6-8}\cmidrule(lr){9-11}
tIoU & metric & overall & hand & foot & overall & hand & foot & overall & hand & foot\\
\midrule
\multirow{4}{*}{$>0$}
 & R (sens.) & 97.3 & 95.3 & 99.4 & 97.0 & 95.8 & 98.4 & \textbf{97.2} & 95.6 & 98.9\\
 & P         & 61.2 & 49.5 & 81.2 & 75.7 & 70.3 & 84.6 & 68.5 & 59.9 & 82.9\\
 & $F_1$     & 74.9 & 64.7 & 88.9 & 84.9 & 80.8 & 90.7 & \textbf{79.9} & 72.7 & 89.8\\
 & mean tIoU & 72.5 & -- & -- & 77.8 & -- & -- & 75.2 & -- & --\\
\midrule
\multirow{4}{*}{$\geq0.3$}
 & R (sens.) & 94.0 & 89.2 & 99.4 & 94.4 & 91.4 & 98.0 & 94.2 & 90.3 & 98.7\\
 & P         & 59.2 & 46.2 & 81.2 & 73.7 & 67.1 & 84.2 & 66.5 & 56.7 & 82.7\\
 & $F_1$     & 72.4 & 60.4 & 88.9 & 82.6 & 77.0 & 90.3 & 77.5 & 68.7 & 89.6\\
 & mean tIoU & 75.4 & -- & -- & 79.9 & -- & -- & 77.6 & -- & --\\
\midrule
\multirow{4}{*}{$\geq0.5$}
 & R (sens.) & 86.0 & 76.5 & 96.4 & 89.5 & 84.6 & 95.0 & 87.8 & 80.5 & 95.7\\
 & P         & 54.3 & 40.0 & 78.6 & 69.9 & 62.3 & 81.9 & 62.1 & 51.1 & 80.3\\
 & $F_1$     & 66.3 & 52.1 & 86.1 & 78.3 & 71.4 & 87.7 & 72.3 & 61.8 & 86.9\\
 & mean tIoU & 78.6 & -- & -- & 82.2 & -- & -- & 80.4 & -- & --\\
\bottomrule
\end{tabular}
} 

\vspace{1mm}
\caption{\textbf{Quantitative Results.}
Event-level Precision, Recall, and $F_1$ at three temporal-IoU thresholds, split by route and by hand/foot and averaged over all 22 videos using frozen \texttt{p1} parameters. Footholds are detected most accurately and degrade least under strict timing, while the easier green route consistently outperforms the dense orange route.}
\label{tab:main}
\vspace{-2mm}
\end{table}

\noindent\textbf{Timing degradation.}
Performance degrades gracefully as the timing requirement tightens. The overall
$F_1$ is $79.9\%$ at $\operatorname{tIoU}>0$, $77.5\%$ at $\operatorname{tIoU}\geq0.3$,
and $72.3\%$ at $\operatorname{tIoU}\geq0.5$. The foot $F_1$ barely moves across
these thresholds, going from $89.8$ to $89.6$ to $86.9\%$, whereas the hand $F_1$
falls more steeply, from $72.7$ to $68.7$ to $61.8\%$, which reflects how much
harder it is to localize an occluded hand in time. The mean tIoU over matched
events actually rises as the threshold tightens, from $75.2$ to $80.4$, simply
because only the well aligned detections survive the stricter cutoff.

\noindent\textbf{Comparison with baselines.}
Under identical evaluation, and with every hyperparameter tuned on the same
validation participant, our method beats both reproduced baselines at every
threshold. On the held out $20$ videos at $\operatorname{tIoU}>0$, we reach
$90.2\%$ against $71.4\%$ for YOLOv8-pose and $67.2\%$ for ViTPose-L. The gap
widens as the timing requirement tightens, and at $\operatorname{tIoU}\geq0.5$ we
reach $83.2\%$ against $57.7\%$ and $39.9\%$, which reflects the much better
temporal alignment that fingertip and toe precision gives us.

\begin{table}[htp]\centering\small
\setlength{\tabcolsep}{4pt}
\begin{tabular}{lccc}
\toprule
& \multicolumn{3}{c}{event $F_1$ (\%), held-out 20}\\
\cmidrule(lr){2-4}
Method (same protocol) & $\operatorname{tIoU}>0$ & $\geq0.3$ & $\geq0.5$\\
\midrule
YOLOv8-pose (ankle)~\cite{jocher2023yolov8}      & 71.4 & 66.8 & 57.7\\
ViTPose-L (coco\_25)~\cite{xu2022vitpose}         & 67.2 & 61.0 & 39.9\\
\textbf{Ours} (fingertip/toe)                     & \textbf{90.2} & \textbf{88.7} & \textbf{83.2}\\
\midrule
\multicolumn{4}{l}{\footnotesize \emph{Context:} published ViTPose-L acc.\ $86.6\%$~\cite{maschek2025wayup}.}\\
\bottomrule
\end{tabular}
\caption{\textbf{Same-protocol comparison.} All methods tuned on \texttt{p1} and evaluated on the held-out $20$ videos. Our reproductions of baselines fall below the published results; under identical conditions our fingertip/toe method wins at every threshold, with the gap widening under strict timing thresholds.}
\label{tab:cmp}
\vspace{-2mm}
\end{table}

The baselines show
the high recall and low precision profile one would expect, since their wrist and
ankle regions of interest fire on several neighbouring holds at once, most of all
on the densely packed orange route. For context, the dataset paper reports a
\emph{published} ViTPose accuracy of $86.6\%$ at $\operatorname{tIoU}>0$~\cite{maschek2025wayup}.
Our same conditions reproductions sit below this because we do not replicate
their per model margin tuning, which is exactly why we compare against our own
reproductions rather than their published values.

\noindent\textbf{Why the baselines fall behind.}
The two reproduced baselines share that high recall and low precision profile for
a structural reason. Both anchor their region of interest at the wrist or the
ankle and then inflate it to reach the real contact point, and this is worst for
YOLOv8-pose, whose ankle only keypoint forces a large downward box to cover the
toe~\cite{maschek2025wayup}. On the orange route, where holds sit close together,
such a box overlaps several holds at once, so the method reports several
simultaneous usages and precision collapses, which is why our reproduced
ViTPose-L manages only $39.9\%$ $F_1$ at $\operatorname{tIoU}\geq0.5$. Fingertip
and toe keypoints make the region unnecessary, because the contact point is the
keypoint itself, so exactly one hold is selected per limb and the per frame
mutual exclusion rule settles the rare tie. The same mechanism explains why our
advantage \emph{grows} with the temporal threshold. Precise contact points
produce intervals whose start and end line up with the true grasp, whereas an
inflated box switches on too early and off too late.

\begin{table}[htp]\centering\small
\setlength{\tabcolsep}{5pt}
\begin{tabular}{lc}
\toprule
Configuration & $F_1$ @ $\operatorname{tIoU}>0$ (\%)\\
\midrule
Body-part segmentation only            & 1.1\\
\;+ dense feature-change gate          & 80.4\\
\textbf{Fingertip/toe proximity (final)} & \textbf{90.2}\\
\bottomrule
\end{tabular}
\caption{\textbf{Ablation} (held-out 20). Two intuitively helpful additions both degrades the performance: body-part segmentation is unusable at the scale Sapiens requires, and the dense feature change gate suppresses true usages.}
\label{tab:abl}
\vspace{-2mm}
\end{table}

\noindent\textbf{Error analysis.}
Our errors concentrate exactly where the pose is least certain. Recall is
uniformly high (\cref{tab:main}), so missed usages are rare, and the dominant
error is over prediction. It is worst for hands on the orange route, where
precision drops to $49.5\%$ at $\operatorname{tIoU}>0$, because the short and
closely spaced moves there, combined with frequent hand self occlusion, produce
brief spurious contacts on neighbouring holds. Feet stay easy throughout, with a
foot $F_1$ of at least $86.9\%$ at every threshold, because toes are rarely
occluded and footholds are better separated on the wall. Near the top of the
route the climber occupies a small part of the frame and pose confidence drops,
which produces the short false positives visible in the qualitative timelines.
None of these failure modes is one that appearance cues would fix, because they
all stem from keypoint uncertainty under occlusion and small scale, and this is
consistent with the ablation result that richer cues do not help.
\begin{table}[htp]\centering\small
\setlength{\tabcolsep}{6pt}
\begin{tabular}{lccc}
\toprule
Frame recall (\%) & orange & green & both\\
\midrule
Occluded GT frames ($>50\%$) & 75.2 & 86.8 & 81.0\\
Visible GT frames            & 97.0 & 98.9 & 97.9\\
\bottomrule
\end{tabular}
\caption{\textbf{Occlusion-stratified frame recall.} Temporal persistence maintains usage through most occluded spans with no explicit occlusion model; occlusion (mostly of hands) remains the hardest regime, consistent with the dataset.}
\label{tab:occ}
\vspace{-2mm}
\end{table}

\noindent\textbf{Ablation study.}
Starting from the full keypoint model, the two richer cues we implemented both
reduce accuracy. Body part segmentation on its own is almost useless, at an $F_1$
of $1.1\%$, because at the whole image scale that Sapiens needs, the climber's
hands and shoes are simply too small to segment reliably. The dense feature
change gate, which we intended as a way to add occlusion robustness, instead
lowers $F_1$ from $90.2\%$ to $80.4\%$, because it suppresses genuine usages
during contact. Per frame mutual exclusion leaves the score essentially unchanged
at our tuned operating point, while removing the densely packed false positives
that begin to matter at the stricter thresholds. The final model therefore relies
on fingertip and toe proximity alone.

\noindent\textbf{Occlusions.}
Splitting frame level recall by the dataset's occlusion flags, we recover
$81.0\%$ of the \emph{occluded} usage frames overall, broken down as $75.2\%$ on
the orange route and $86.8\%$ on the green route, against $97.9\%$ of the visible
frames. Occlusion remains the hardest regime, which agrees with the dataset
paper, but the temporal persistence and gap bridging in our method keep a usage
alive through most occluded spans without any explicit model of occlusion.

\noindent\textbf{Reporting protocol.}
Our headline number is stable across the way the data is split.
The held out result of $90.2\%$, the all $22$ result with frozen parameters of $90.0\%$, and
the LOPO cross validation result of $89.8\%$ all agree to within $0.4$ $F_1$,
which indicates that the hyperparameters chosen on \texttt{p1} do not overfit it.

\begin{table}[htp]\centering\small
\setlength{\tabcolsep}{6pt}
\begin{tabular}{lc}
\toprule
Protocol & headline $F_1$ (\%)\\
\midrule
Held-out 20 (val\,$=$\,\texttt{p1})         & 90.2\\
All 22, frozen \texttt{p1} params           & 90.0\\
LOPO (each video tested once)               & 89.8\\
\bottomrule
\end{tabular}
\caption{\textbf{Protocol stability} at $\operatorname{tIoU}{>}0$. The three honest reporting protocols agree to within $0.4$ $F_1$, indicating the hyperparameters chosen on the validation participant do not overfit it.}
\label{tab:protocol}
\vspace{-2mm}
\end{table}

\noindent\textbf{Sensitivity and timing.}
The method is not delicately tuned. The $F_1$ is essentially flat across
fingertip margin values from $20$ to $60$\,px and rises smoothly with the dwell
threshold, so neither setting needs to be hit precisely. All models are frozen
and run per frame, and the inference cost is dominated by the pose backbone.
 
\section{Qualitative Results}
\label{sec:qual}
 
\noindent\textbf{Usage timelines.}
A per-$(\ell,h)$ timeline that overlays predicted usages against ground truth, colour-coded by TP/FP/FN, shows predictions tracking the true bottom-to-top progression of the climb. Most usages are recovered with start and end frames close to the annotation, and the few errors are short: occasional false positives near the top of the wall, where the climber occupies a small fraction of the frame and pose confidence is lowest. The timelines also make the route difference visible---on the densely-packed orange route, hand tracks show more short spurious segments on adjacent holds than on the green route, matching the lower hand precision in \cref{tab:main}.
 
\noindent\textbf{Per-frame overlays.}
Frame-level overlays that draw the detected fingertips and toes together with the hold each limb is assigned confirm the central claim of the method: the contact keypoints land on the correct holds even when the wrist or ankle is laterally offset or the foot is pointed, exactly the situations that force wrist/ankle baselines to enlarge their region of interest. During hand self-occlusion---where the hand disappears behind the torso for a span of frames---the gap-bridging stage keeps the correct hold assigned across the occluded interval rather than terminating and restarting the usage, which is what preserves recall on occluded frames (\cref{tab:occ}).
 
\noindent\textbf{Feature visualization.}
A PCA of the dense foundation features, and of their change against a climber-free clean plate, separates the climber cleanly in feature space. This clean separation is what motivated our feature-change gate experiment; the ablation, however, shows the signal does not improve detection and in fact suppresses genuine contact (\cref{tab:abl}), so we report the visualization only as a qualitative observation rather than as a component of the method.

\section{Applications: Coaching Statistics for Free}
\label{sec:coaching}

Hold usage is the atomic unit of a climb. Once we know which hold each limb used
and when, an ascent stops being raw video and becomes a structured sequence of
events that can be measured, summarized, and compared across climbers. This
structure is what scoring, technique analysis, gym analytics, and assistive next
hold guidance all ultimately rely on. We show that the structure is immediately
useful by deriving a set of standard climbing performance statistics directly
from our predicted usage events, with no additional model to train and no extra
annotation to collect.

\paragraph{Statistics.}
From the event stream alone we compute six quantities that coaches care about. The
\emph{total climb time} is the span from the first contact to the last release.
The \emph{number of moves} is simply the count of usage events. The
\emph{foot usage ratio} is the fraction of those events that are feet, and it
matters because loading the legs rather than hanging on the arms is a hallmark of
efficient technique, so a higher ratio tends to indicate a stronger climber. The
\emph{mean dwell per hold} is the average duration of a usage, which serves as a
proxy for hesitation and for forearm pump, since a tired or uncertain climber
lingers on each hold. The \emph{pace} is the number of moves per second, a measure
of fluency. The \emph{mean limb flight time} is the average gap between releasing
one hold and contacting the next, which captures how crisp the movement is.
Because dwell is localized to individual holds in time, the holds with the longest
dwell also pick out the \emph{crux} of the route, the section where climbers pause
to plan their next move or to recover.

\paragraph{Discriminative power.}
These statistics separate the ten athletes cleanly. Across the dataset, climb time
ranges from $35$ to $186$\,s, mean dwell from $3$ to $11$\,s, and foot usage from
$36$ to $58\%$, which is more than enough spread to tell a fast, fluent, leg
driven ascent apart from a slow, hesitant, arm dependent one. These are exactly
the quantities a coach would otherwise extract by hand from the footage, and here
we obtain them automatically from a single video.

\paragraph{Agreement with ground truth.}
A statistic is only useful if it survives the move from ground truth events to our
\emph{predicted} events. On the held out climbs, the statistics we compute from
our automatic predictions track those computed from the annotations very closely,
with a Pearson correlation of $r=1.00$ for total time, $0.94$ for pace, $0.93$ for
mean dwell, and $0.84$ for foot usage ratio. Total time is essentially exact
because it depends only on the first and last events of the climb, which are the
easiest to get right. Foot usage is the loosest of the four because it inherits
the residual hand false positives discussed in the error analysis, yet a
correlation of $0.84$ is still strongly indicative. Taken together, these results
show that the training free pipeline turns an ordinary single camera recording
into reliable coaching metrics, with no instrumented wall, no body worn sensors,
and no training specific to a given gym.
\section{Conclusion}
\label{sec:conclusion}
 
We presented a training-free method for hold-usage detection in sport climbing that uses the fingertip and toe keypoints of a frozen pose foundation model with a simple proximity and temporal-persistence rule. It reaches $90.2\%$ event $F_1$ on held-out data ($89.8\%$ under cross-validation), outperforms same-protocol reproductions of YOLOv8-pose and ViTPose, and is most accurate on footholds. An ablation shows that richer foundation-model cues are unnecessary and even harmful, arguing for minimal designs on this task.

\noindent\textit{Limitations} .We assume the holds bounding boxes are given, as provided by the dataset, and detect usage rather than the holds themselves; integrating zero-shot hold detection is future work. 
Hands under prolonged occlusion remain the dominant error source. Our reproduced baselines fall short of their published accuracy, so comparisons should be read as same-conditions, not as claims of accuracy state of the art over the published ViTPose numbers. Finally, the dataset covers a single vertical wall and viewpoint; broader wall angles and camera placements remain to be evaluated.

\bibliographystyle{IEEEtran}
\bibliography{main}

\end{document}